\documentclass[letterpaper, 10 pt, conference]{ieeeconf_old}
\IEEEoverridecommandlockouts    
\usepackage{graphics}           
\usepackage{times}              
\usepackage{amsmath}            
\usepackage{amssymb}            
\usepackage{graphicx}
\usepackage{algorithm}
\usepackage[noend]{algpseudocode}
\usepackage{booktabs}
\usepackage{color}
\definecolor{instructioncolor}{rgb}{.5,.5,.5}

\usepackage[font=small]{caption}

\def\figref#1{Fig.~\ref{#1}}

\def\eqref#1{Eq.~(\ref{#1})}

\makeatletter
\usepackage{xspace}
\DeclareRobustCommand\onedot{\futurelet\@let@token\@onedot}
\def\@onedot{\ifx\@let@token.\else.\null\fi\xspace}

\makeatother

\usepackage{array}
\newcolumntype{L}[1]{>{\raggedright\let\newline\\\arraybackslash\hspace{0pt}}m{#1}}
\newcolumntype{C}[1]{>{\centering\let\newline\\\arraybackslash\hspace{0pt}}m{#1}}
\newcolumntype{R}[1]{>{\raggedleft\let\newline\\\arraybackslash\hspace{0pt}}m{#1}}

\usepackage{multirow}
\usepackage{hyperref}
\usepackage[table]{xcolor}
\usepackage{placeins}  
\usepackage{cite}
\usepackage{booktabs}
\usepackage{amsmath}
\usepackage{amssymb}
\usepackage{svg}
\usepackage{graphicx}

\title{\LARGE \bf 
SPRINT: Efficient Spectral Priors for Humanoid Athletic Sprints}

\author{Yantong Wei$^{1,\dagger}$, Kaihong Huang$^{1,\dagger}$, Hainan Pan$^{1}$, Jiawei Luo$^{1}$, Jiawei Zhou$^{1}$,\\%
Ziyan Mai$^{1}$, Zhiwen Zeng$^{1}$, Yaonan Wang$^{2}$, Huimin Lu$^{1,*}$
\thanks{$^1$Y. Wei, K. Huang, H. Pan, J. Luo, J. Zhou, Z. Mai, Z. Zeng and H. Lu are with the College of Intelligence Science and Technology, National University of Defense Technology, China.}
\thanks{$^2$Y. Wang is with the School of Artificial Intelligence and Robotics, Hunan University, China.}
\thanks{$^{\dagger}$Co-first author, $^{*}$Corresponding author.}
}

\begin{document}
\maketitle
\thispagestyle{empty}
\pagestyle{empty}
\begin{abstract}


The pursuit of humanoid athletic sprints is hindered by a scarcity of humanoid-viable kinematic reference data and the inability of existing frameworks to maintain stability during sprints. To overcome these limitations, we introduce SPRINT, a novel framework driven by efficient, frequency-adaptive spectral priors. By characterizing the fundamental periodicity of human locomotion in the frequency domain using a reference library of five discrete motion sequences, these priors generate kinematically feasible joint trajectories across a broad velocity spectrum, successfully extrapolating to speeds that exceed the reference distribution. Guided by these pretrained priors, the SPRINT policy achieves zero-shot sim-to-real transfer in field experiments on the Unitree G1 platform, reaching a peak sprinting velocity of 6\,m/s and demonstrating seamless gait transitions while preserving biomimetic naturalness. Ultimately, this work establishes frequency-adaptive spectral priors as a highly data-efficient foundation for humanoid athletic sprints. The project page is available at \url{https://anonymous.4open.science/w/SPRINT-138A/}.

\end{abstract}
\vspace{0.5em}
\section{Introduction}
\label{sec:intro}

\PARstart{A}{thletic} sprinting represents a critical frontier in humanoid high-speed locomotion. However, realizing stable, high-velocity performance is hindered by a scarcity of kinematic reference data viable for humanoid robots. Furthermore, training robust control policies is complicated by the dual necessity of maintaining tracking precision during high dynamic sprints and executing seamless gait transitions across a broad velocity spectrum.

Reinforcement Learning (RL) is widely utilized to train complex robotic skills \cite{radosavovic2024real,li2025reinforcement}. While reward-based optimization facilitates basic task completion, objective-driven training yields feasible locomotion patterns that lack biomimetic naturalness. Adversarial Motion Priors (AMP) mitigate this limitation by motivating robots to mimic human demonstrations \cite{zhang2024whole,tang2024humanmimic}. However, adversarial frameworks are prone to training instability and mode collapse during high-speed sprints, due to the scarcity of corresponding kinematic reference data \cite{arjovsky2017wasserstein}. Similarly, alternative imitation learning approaches \cite{he2025asap,peng2018deepmimic,chen2025gmt,liao2508beyondmimic} achieve robust tracking in highly dynamic scenarios but remain limited in their capacity to support continuous velocity scaling and seamless gait transitions. Furthermore, hybrid strategies, such as Adaptive Imitated Central Pattern Generators (AI-CPG) \cite{li2024ai}, remain restricted to peak speeds below 4\,m/s in simulation, leaving a critical gap in physical, highly dynamic athletic performance.

\begin{figure}[t] 
\centering 
\includegraphics[width=\columnwidth]{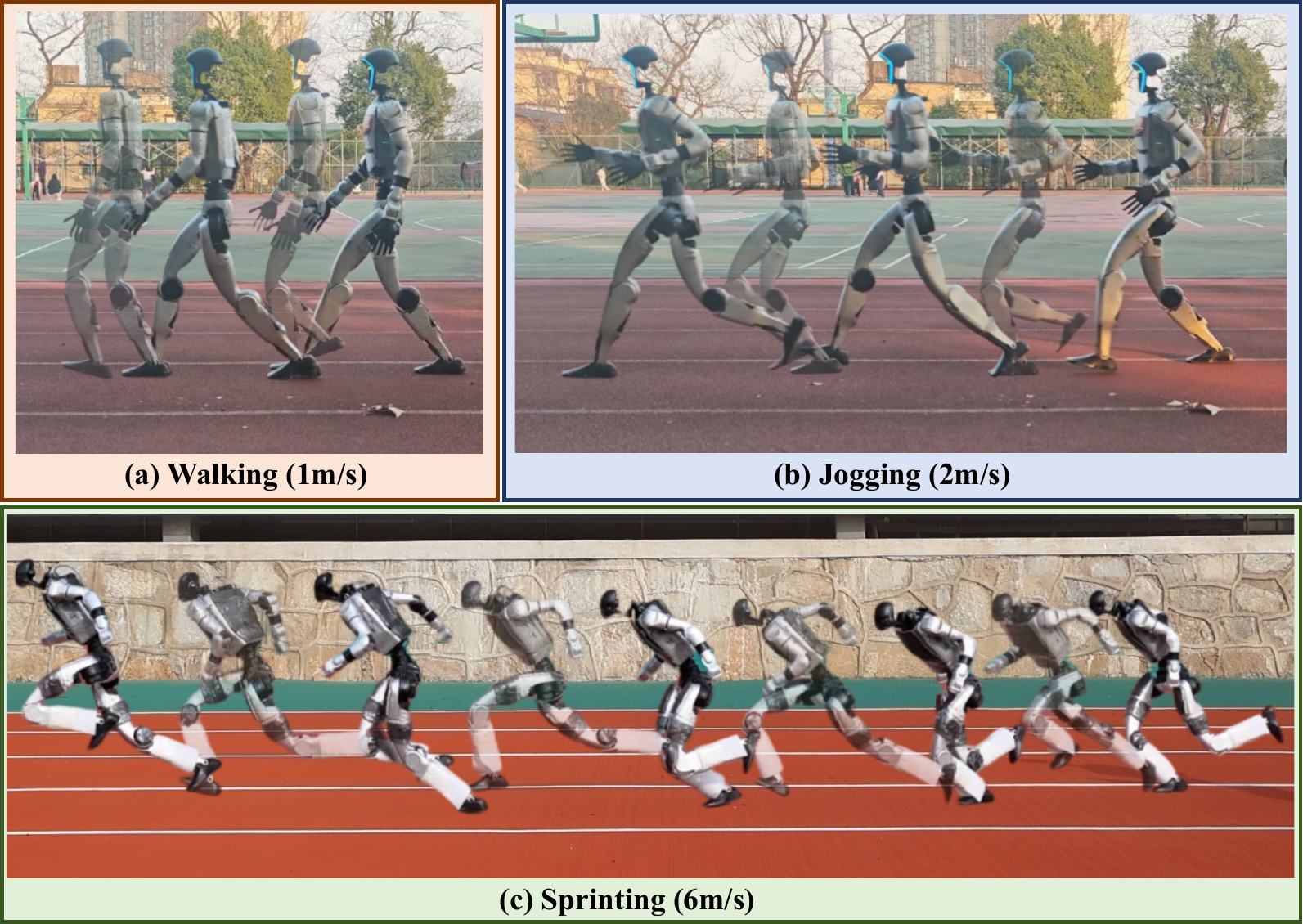}
\caption{Versatile humanoid locomotion. SPRINT supports a wide range of natural gaits, including walking, jogging, and sprinting.}
\vspace{-0.2cm} 
\label{fig:motivation} 
\end{figure}


Frequency-adaptive spectral priors form the core of the SPRINT framework, providing fine-grained kinematic guidance to ensure authentic athletic form during high-speed locomotion. We train these spectral priors using a reference library of only five motion sequences (428 frames, 14\,s) comprising discrete walking, jogging, and running gaits. By characterizing the inherent periodicity of human locomotion in the frequency domain, we effectively overcome data scarcity. The spectral priors map rhythmic features into kinematically feasible joint trajectories across a broad velocity spectrum, successfully extrapolating to speeds that exceed the reference distribution.


The SPRINT framework employs a hierarchical architecture integrating high-level spectral priors with low-level stabilization. The pretrained frequency-adaptive spectral priors generate joint trajectories that serve as the primary kinematic reference, while the low-level stabilization employs residual actions to compensate for unmodeled system dynamics and environmental disturbances. In simulation, the SPRINT policy achieves a peak sprinting velocity of 6\,m/s while preserving high biomimetic naturalness. Through dynamics randomization and observation noise modeling, we successfully execute zero-shot sim-to-real transfer to the physical Unitree G1 platform (\figref{fig:motivation}). Ultimately, extensive field experiments validate the capability of the SPRINT policy to execute versatile humanoid locomotion and seamless gait transitions across a broad velocity spectrum.


Our contributions are summarized as follows:
\begin{itemize}
\item \textbf{Frequency-Adaptive Spectral Priors}: We introduce efficient spectral priors capable of generating kinematically feasible joint trajectories for high-speed sprints.
\item \textbf{SPRINT Policy}: By integrating the pretrained priors with residual actions, we achieve robust humanoid locomotion and enable seamless gait transitions across a velocity spectrum of 0 to 6\,m/s.
\item \textbf{Extensive Field Validation}: We demonstrate zero-shot sim-to-real transfer on the Unitree G1 platform, verifying the framework's capability to execute high-speed sprints while preserving biomimetic naturalness.
\end{itemize}

\begin{figure*}[htbp]
\centering

\includegraphics[width=\textwidth]{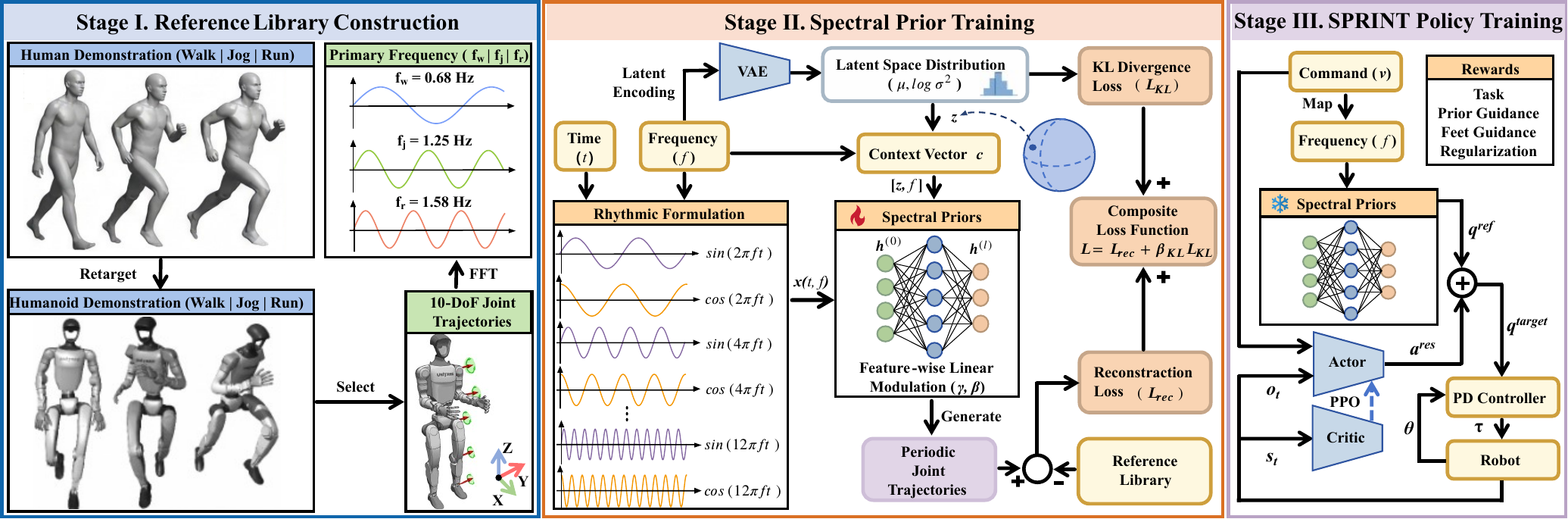}
\caption{Three-stage workflow of the SPRINT framework. In \textbf{Stage I}, rhythmic features are extracted from retargeted humanoid demonstrations via spectral analysis. In \textbf{Stage II}, these rhythmic features form the basis for training the frequency-adaptive spectral priors. In \textbf{Stage III}, a hierarchical control structure is employed, coupling high-level spectral priors with low-level stabilization.}

\vspace{-0.2cm}
\label{fig:pipline}
\end{figure*}

\section{Related work}

\subsection{Learning-based Humanoid Locomotion}

Reinforcement learning (RL) has emerged as a leading paradigm for robotic locomotion, obviating the heavy reliance on precise analytical dynamics models traditionally required by classical methods \cite{herdt2010online}. For instance, RL has empowered the bipedal robot Cassie to master high-dimensional, nonlinear dynamics during high-speed running at 5 m/s \cite{crowley2023optimizing}. Furthermore, recent advances in full-body humanoid control leverage frameworks like Humanoid-Gym \cite{gu2024humanoid} to enhance environmental adaptability and locomotion robustness. However, these strategies often rely on manual reward engineering, which yields feasible but unnatural movements, as the optimization process prioritizes basic task completion over biomimetic naturalness. To overcome this inherent lack of biomimetic naturalness, we introduce frequency-adaptive spectral priors that provide fine-grained kinematic guidance to ensure authentic athletic form during high-speed locomotion.

\subsection{Data-driven Motion Generation}

Data-driven motion generation in computer animation employs Variational Autoencoders (VAEs) \cite{guo2020action2motion}, Transformers \cite{petrovich2022temos}, and diffusion models \cite{tevet2023human} to synthesize humanoid pose sequences from text or video. However, these sequences are kinematically infeasible for direct robotic control due to morphological and physical constraints. In robotics, generative architectures such as the Generative Motion Prior (GMP) \cite{zhang2025natural} guide control policies by predicting future trajectories from current states. The Conditional Motion Generator (CMG) \cite{li2025run} achieves natural walking and running at speeds up to 2.5\,m/s on the Unitree G1 platform, and the Adaptive Imitated Central Pattern Generator (AI-CPG) \cite{li2024ai} bridges distinct gaits via frequency-domain analysis. Despite facilitating rhythmic feedforward patterns, AI-CPG remains confined to idealized simulations and peak speeds below 4\,m/s. Bound by fixed reference trajectories, these models fail to extrapolate to the dynamic regimes of athletic sprinting. To overcome these limitations, we formulate a highly efficient, frequency-conditioned latent space to train our spectral priors, enabling robust velocity extrapolation far beyond the reference distribution.

\subsection{Imitation from Human Motion}

Integrating authentic athletic form into robotic controllers requires high-fidelity kinematic reference data directly applicable to humanoid robots. Adversarial Motion Priors (AMP) \cite{goodfellow2014generative,peng2021amp} address this challenge by treating retargeted human demonstrations as a target distribution for policy guidance. While AMP achieves locomotion with high biomimetic naturalness, it is prone to training instability in high-speed scenarios due to the scarcity of sprinting reference data \cite{arjovsky2017wasserstein}. This inherent data bottleneck has motivated the development of alternative imitation learning frameworks focused on robust motion tracking \cite{he2025asap,peng2018deepmimic,chen2025gmt,liao2508beyondmimic}. Although these specialized frameworks successfully execute highly dynamic motions, they remain constrained to predefined kinematic trajectories, lacking the adaptability required for continuous velocity transitions. The SPRINT framework resolves these limitations by integrating frequency-adaptive spectral priors with residual actions, simultaneously achieving athletic agility and robust stability across a continuous velocity spectrum of 0 to 6\,m/s.

\section{Methodology}



As illustrated in \figref{fig:pipline}, the SPRINT framework consists of three primary stages: reference library construction, spectral prior training, and SPRINT policy training. Initially, we utilize spectral analysis to extract rhythmic features from the curated reference library. These features subsequently form the foundation for training the frequency-adaptive spectral priors. During policy training, these pretrained priors generate the primary kinematic reference trajectories, while residual actions compensate for environmental disturbances.

\subsection{Reference Library Construction (Stage I)}


The objective of Stage I is to construct a highly curated, kinematically feasible reference library that provides the inductive bias required to train the spectral priors. Initially, we extract discrete human locomotion patterns from the LAFAN1 dataset \cite{harvey2020robust}. To bridge the morphological gap between these human demonstrations and the target robot, we design a comprehensive four-step curation pipeline:

\textbf{\textit{1) Morphological Retargeting}}.
We leverage the GMR framework \cite{araujo2025retargeting} to bridge the morphological discrepancies between human demonstrations and target humanoid robots.

\textbf{\textit{2) Joint Selection}}.
We isolate a 10-DoF joint subset $\mathcal{J}$, to efficiently model rhythmic dynamics. This subset comprises the hip pitch, knee, and ankle pitch joints for lower-limb propulsion, alongside the shoulder pitch and elbow joints to capture the upper-limb swing necessary for balance.

\textbf{\textit{3) Data Normalization}}.
We extract five single-cycle gait sequences, standardizing each to a 10\,s duration. Subsequently, we apply Fourier resampling to increase the temporal resolution from 30\,Hz to 60\,Hz, followed by a Savitzky-Golay filter to eliminate high-frequency kinematic noise.

\textbf{\textit{4) Spectral Analysis}}.
We employ a Fast Fourier Transform (FFT) to compute the Power Spectral Density (PSD) for each joint in $\mathcal{J}$. This analysis identifies the primary frequencies (e.g., $f_w$, $f_j$, $f_r$) and reveals a half-cycle phase offset with nearly identical amplitudes between contralateral limbs.


Collectively, this curation pipeline yields a highly representative reference library of periodic 10-DoF joint trajectories. Comprising exactly five sequences that encompass walking, jogging, and running gaits, this reference library establishes a precise kinematic mapping between velocities ($\{0.66, 1.10, 2.29, 2.87, 3.40\}$\,m/s) and primary frequencies ($\{0.68, 0.86, 1.25, 1.36, 1.58\}$\,Hz). Ultimately, this curated reference library provides the essential inductive bias required to train the frequency-adaptive spectral priors.

\subsection{Spectral Prior Training (Stage II)}


The objective of Stage II is to train frequency-adaptive spectral priors capable of generating kinematically feasible joint trajectories across a continuous velocity spectrum. To achieve this, we formulate a neural architecture that maps temporal phase structures and target frequencies directly to high-fidelity joint trajectories. The training pipeline for the spectral priors is structured into four sequential phases:


\textbf{\textit{1) Frequency-Conditioned Latent Encoding}}.
To capture the gait styles associated with varying speeds, we employ a Variational Autoencoder (VAE) \cite{won2022physics} to map the target frequency $f$ into a continuous latent space. The encoder network predicts the parameters of a multivariate Gaussian distribution, specifically the mean vector $\boldsymbol{\mu}$ and the log-variance vector $\log\boldsymbol{\sigma}^2$, parameterizing the conditional posterior distribution $q(\boldsymbol{z}|f) = \mathcal{N}(\boldsymbol{\mu}, \text{diag}(\boldsymbol{\sigma}^2))$. To ensure differentiability during backpropagation, we apply the standard reparameterization trick to sample the latent style vector $\boldsymbol{z}$:
\begin{equation}
\boldsymbol{z} = \boldsymbol{\mu} + \boldsymbol{\sigma} \odot \boldsymbol{\epsilon},
\end{equation}
where $\boldsymbol{\epsilon} \sim \mathcal{N}(\boldsymbol{0}, \boldsymbol{I})$ represents standard Gaussian noise, and $\odot$ denotes the element-wise product.


\textbf{\textit{2) Multi-Harmonic Rhythmic Formulation}}.
To capture the periodicity inherent in human locomotion, we formalize the temporal gait structure as a multi-harmonic vector. Given a continuous time variable $t \in [0, 10]$\,s and a target frequency $f$, we define the instantaneous phase as $\phi = 2\pi ft$. To accurately model complex, non-sinusoidal joint trajectories, we expand this phase into $K=6$ harmonic components, yielding the spectral vector $\boldsymbol{x}(t, f) \in \mathbb{R}^{2K}$:
\begin{equation}
\boldsymbol{x}(t,f) = \begin{bmatrix} \sin(\phi) & \cos(\phi) & \cdots & \sin(K\phi) & \cos(K\phi) \end{bmatrix}^{\top},
\end{equation}
This multi-harmonic vector serves as the initial input state $\boldsymbol{h}^{(0)}$ for the generative decoder network.

\textbf{\textit{3) Context-Modulated Trajectory Generation}}.
We achieve trajectory generation by fusing the rhythmic input with the sampled latent style. Initially, a context vector $\boldsymbol{c} = [\boldsymbol{z}, f]$ is formed by concatenating the latent vector and the frequency. This context vector drives a Feature-wise Linear Modulation (FiLM) generator \cite{perez2018film} to compute scaling factors $\boldsymbol{\gamma}^{(l)}$ and offsets $\boldsymbol{\beta}^{(l)}$ for each hidden layer $l$ of the trajectory decoder. To guarantee numerical stability and prevent signal saturation across deep layers, we bound these modulation parameters using hyperbolic tangent functions:
\begin{equation}
\boldsymbol{\gamma}^{(l)} = 1 + 0.1\tanh(\boldsymbol{W}_\gamma^{(l)}\boldsymbol{c} + \boldsymbol{b}_\gamma^{(l)}),
\end{equation}
\begin{equation}
\boldsymbol{\beta}^{(l)} = 0.1\tanh(\boldsymbol{W}_\beta^{(l)}\boldsymbol{c} + \boldsymbol{b}_\beta^{(l)}),
\end{equation}
where $\boldsymbol{W}_\gamma^{(l)}, \boldsymbol{W}_\beta^{(l)}, \boldsymbol{b}_\gamma^{(l)}$, and $\boldsymbol{b}_\beta^{(l)}$ denote the layer-specific trainable weight matrices and bias vectors of the FiLM network. The decoder sequentially processes the initial multi-harmonic input $\boldsymbol{h}^{(0)}$ through hidden layers equipped with SiLU activations ($\sigma$). At each layer $l$, the corresponding FiLM parameters linearly modulate the intermediate features:
\begin{equation}
\boldsymbol{h}^{(l)} = \sigma( \boldsymbol{\gamma}^{(l)} \odot \boldsymbol{W}^{(l)}\boldsymbol{h}^{(l-1)} + \boldsymbol{\beta}^{(l)} ),
\end{equation}
where $\boldsymbol{W}^{(l)}$ represents the standard layer weights. This modulated forward pass maps the periodic signals into high-fidelity trajectories for the 10-DoF joint subset $\mathcal{J}$.

\textbf{\textit{4) Composite Optimization}}.
To ensure kinematic tracking while maintaining a well-structured latent space, we optimize the architecture using a composite loss function: $L = L_{\text{rec}} + \beta_{\text{KL}}L_{\text{KL}}$. The reconstruction loss ($L_{\text{rec}}$) evaluates the generator's performance by computing the Mean Squared Error (MSE) between the generated trajectories and the reference library. Simultaneously, the Kullback-Leibler (KL) divergence loss ($L_{\text{KL}}$), weighted by the hyperparameter $\beta_{\text{KL}}$, regularizes the predicted distribution parameters ($\boldsymbol{\mu}$ and $\log\boldsymbol{\sigma}^2$) toward a standard normal prior $\mathcal{N}(\boldsymbol{0}, \boldsymbol{I})$. We isolate the structural regularization of the latent space, thereby ensuring robust convergence during spectral prior training.

\subsection{SPRINT Policy Training (Stage III)}


The objective of Stage III is to train a policy capable of executing high-speed sprints while preserving the biomimetic naturalness established by the pretrained spectral priors. To overcome the unnatural locomotion patterns associated with objective-driven RL, we formulate the high-speed locomotion problem as a Partially Observable Markov Decision Process (POMDP). We solve this POMDP utilizing a hierarchical residual control architecture optimized via Proximal Policy Optimization (PPO) \cite{schulman2017proximal}. The training pipeline for this policy is structured into three sequential phases:

\textbf{\textit{1) SPRINT Policy Formulation}}.
To maximize the expected cumulative discounted reward $J(\theta) = \mathbb{E}_{\tau \sim \pi_{\theta}}\left[ \sum_{t=0}^{T} \gamma^{t} r_{t} \right]$, the SPRINT policy decouples kinematic generation from dynamic stabilization. We utilize the frozen, pretrained spectral priors as a robust kinematic baseline. These priors generate biomimetic reference joint trajectories, denoted as $\boldsymbol{q}^{\text{ref}}$. Consequently, the RL policy $\pi_\theta$ outputs a residual action $\boldsymbol{a}^{\text{res}}$, modulated by a scalar gain $\alpha$, to compensate for unmodeled system dynamics. The final target joint position tracked by the PD controller is defined as:
\begin{equation}
\boldsymbol{q}^{\text{target}} = \boldsymbol{q}^{\text{ref}} + \alpha \boldsymbol{a}^{\text{res}}.
\end{equation}
This hierarchical structure leverages the spectral priors, thereby efficiently constraining the RL exploration space to focus exclusively on reactive stabilization.


\begin{table}[htbp]
\centering
\caption{Definitions of Feet Guidance and Regularization Rewards}
\label{tab:combined_rewards}
\renewcommand{\arraystretch}{1.5}
\begin{tabular*}{\columnwidth}{@{\extracolsep{\fill}} lcc}
\toprule
\textbf{Reward Term} & \textbf{Definition} & \textbf{Weights} \\
\midrule
Close Feet Penalty   & $\max(0, d_{\min} - |y_{l} - y_{r}|)$ & $-100.0$ \\
Feet Air Height      & $\sum \mathbb{I}(\text{air}) \cdot \exp\left(-\frac{|h_i - h_{\text{ref}}|}{\sigma}\right)$ & $5.0$ \\
Low Speed Air        & $\mathbb{I}(v_t < 1.5) \cdot \mathbb{I}(\text{both air})$ & $-6.0$ \\
High Speed Ground    & $\mathbb{I}(v_t > 3.0) \cdot \mathbb{I}(\text{both ground})$ & $-3.0$ \\
Feet Ground Parallel & $\sum \exp(-\|\boldsymbol{g}_{xy}^{\text{foot}}\|) \cdot \mathbb{I}(\text{contact})$ & $1.0$ \\ 
Feet Slide Penalty   & $\sum \|\boldsymbol{v}_{\text{feet}}^{\text{tangent}}\| \cdot \mathbb{I}(\text{contact})$ & $-10.0$ \\ 
Alive Bonus          & $1 - \mathbb{I}(\text{terminated})$ & $1.0$ \\ 
Torque Penalty       & $\sum \|\boldsymbol{\tau}_i\|^2$ & $-5 \times 10^{-6}$ \\
Joint Accel Penalty  & $\sum \|\ddot{\boldsymbol{q}}_i\|^2$ & $-2 \times 10^{-8}$ \\
Joint Limits Penalty & $\sum \mathbb{I} (q_i \notin [q_{\min}, q_{\max}])$ & $-10.0$ \\
Joint Vel Penalty    & $\sum \|\dot{\boldsymbol{q}}_i\|^2$ & $-5 \times 10^{-4}$ \\
R-P Ang Vel Penalty  & $\|\boldsymbol{\omega}_{xy}\|^2$ & $-0.5$ \\
Base Height Penalty  & $\|z_{\text{base}} - z_{\text{target}}\|^2$ & $-30.0$ \\
\bottomrule
\end{tabular*}
\end{table}

\textbf{\textit{2) Reward Design}}.
A composite reward function $r$ is designed to balance task tracking precision, biomimetic naturalness, and hardware safety constraints:
\begin{equation}
r = r_{\text{task}} + r_{\text{prior}} + r_{\text{feet}} + r_{\text{reg}}.
\end{equation}

\begin{itemize}
\item \textbf{Task Reward} ($r_{\text{task}}=r_{v}+r_{\omega}$): To mitigate destabilizing transients during rapid acceleration, the linear velocity reward $r_{v}$ tracks a smoothly filtered command $\tilde{v}_{x, t}^{\text{cmd}}$, bounded by an acceleration limit $a$:
\begin{equation}
\tilde{v}_{x, t}^{\text{cmd}} = \tilde{v}_{x, t-1}^{\text{cmd}} + \text{clip}(v_{x}^{\text{cmd}} - \tilde{v}_{x, t-1}^{\text{cmd}}, -a\Delta t, a\Delta t).
\end{equation}
The tracking precision for both the filtered linear velocity $\tilde{v}_{x, t}^{\text{cmd}}$ and the raw angular velocity $\boldsymbol{\omega}_{z}^{\text{cmd}}$ is evaluated using exponential kernels:
\begin{equation}
r_{v} = w_v \exp \left( - \frac{\| \tilde{\boldsymbol{v}}_{x, t}^{\text{cmd}} - \boldsymbol{v} \|^2}{\sigma_v} \right),
\end{equation}

\begin{equation}
r_{\omega} = w_{\omega} \exp \left( - \frac{\| \boldsymbol{\omega}_{z}^{\text{cmd}} - \boldsymbol{\omega} \|^2}{\sigma_{\omega}} \right).
\end{equation}
\end{itemize}

\begin{itemize}
\item \textbf{Prior Guidance Reward} ($r_{\text{prior}}$): To ensure the policy adheres to the human-like gait, this term penalizes deviations from the baseline spectral prior trajectories $\boldsymbol{q}_j^{\text{ref}}$ across the 10-DoF joint subset $\mathcal{J}$:

\begin{equation}
r_{\text{prior}} = w_p \sum_{j \in \mathcal{J}} \exp \left( - \frac{\|\boldsymbol{q}_j - \boldsymbol{q}_j^{\text{ref}}\|^2}{\sigma_q} \right).
\end{equation}
\end{itemize}

\begin{itemize}
\item \textbf{Feet Guidance Reward} ($r_{\text{feet}}$): High-speed locomotion requires precise regulation of the stance and swing phases. We enforce a velocity-dependent minimum flight time target $t_{\text{target}}$, derived from the command frequency $f_{\text{cmd}}$ and the required swing phase duty cycle:

\begin{equation}
t_{\text{target}} = \frac{D_{\text{swing}}}{f_{\text{cmd}}},
\end{equation}
where $D_{\text{swing}} = 0.4$ represents the predefined proportion of the swing phase relative to the total gait cycle. By comparing the accumulated actual air time $t_{\text{air}}$ to this target, the reward encourages sufficient swing duration for solving complex ground interactions:

\begin{equation}
r_{\text{a}} = w_{\text{a}}\sum \mathbb{I}(\text{air}) \max(0, t_{\text{target}} - t_{\text{air}}).
\end{equation}

\end{itemize}
\begin{itemize}
\item \textbf{Regularization Reward} ($r_{\text{reg}}$): Maintaining equilibrium during sprints requires a forward shift of the center of mass. To induce this behavior, the torso pitch penalty adaptively constrains the projected gravity component $g_{x}$. It establishes a speed-dependent forward tilt target $g_{f} = 0.1 + 0.05 v_{x}^{\text{cmd}}$ to facilitate propulsion and dynamic balance, while penalizing any backward tilt ($g_{b} = 0$):
\begin{equation}
r_{\text{t}} = w_{\text{t}}\left( \max(0, g_{x} - g_{f}) + \max(0, g_{b} - g_{x}) \right)^2.
\end{equation}
\end{itemize}

Comprehensive details of all supplementary feet guidance and regularization terms are summarized in Table \ref{tab:combined_rewards}.

\textbf{\textit{3) Sim-to-Real Transfer Strategy}}.
To bridge the sim-to-real gap, we integrate an Asymmetric Actor-Critic (AAC) architecture \cite{pinto2018asymmetric}, progressive curriculum learning, and dynamics randomization. The AAC framework facilitates policy optimization through structured informational asymmetry. The actor network, tailored for real-time onboard deployment, processes exclusively a proprioceptive observation vector $\boldsymbol{o}_t$ (comprising angular velocity, projected gravity, task commands, joint kinematics, and the previous residual action $\boldsymbol{a}_{t-1}^{\text{res}}$). Conversely, the critic network evaluates a privileged state vector $\boldsymbol{s}_t$, which augments $\boldsymbol{o}_t$ with the linear velocity and the reference trajectories generated by the spectral priors. Concurrently, a progressive curriculum learning scheme gradually scales target velocity commands, while dynamics randomization perturbs key physical properties (e.g., mass, ground friction, and motor latency). Collectively, these techniques profoundly enhance the policy's resilience to environmental disturbances, ensuring the physical hardware deployment strictly preserves the biomimetic naturalness.

\begin{figure*}[htbp]
\centering

\includegraphics[width=\textwidth]{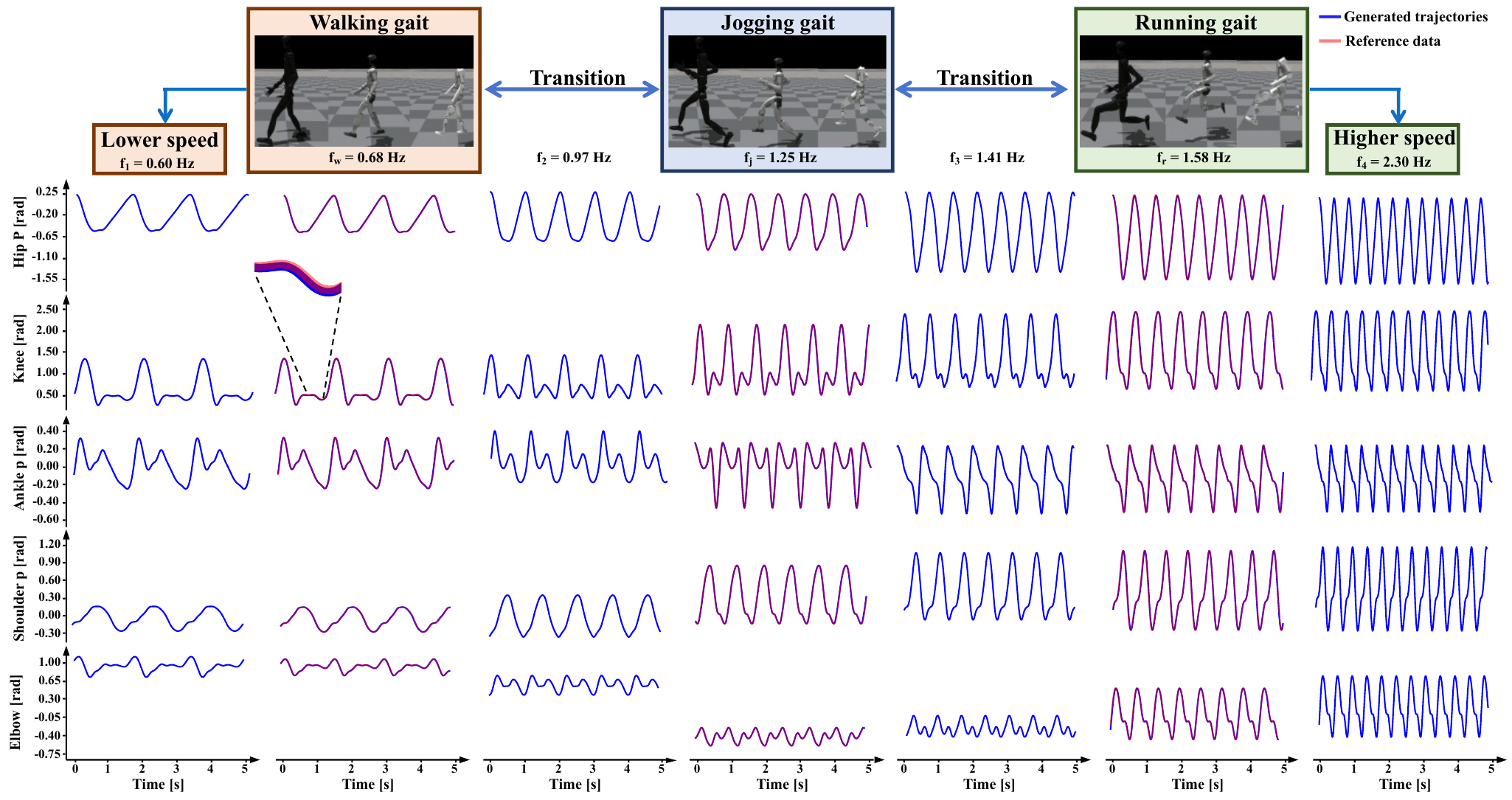}
\caption{Evaluation of frequency-adaptive spectral priors. \textbf{First row}: Morphological adaptability enables seamless gait synthesis across diverse humanoid scales. \textbf{Second row}: Generated trajectories (blue) precisely align with discrete reference data (red) at the predefined reference frequencies ($f_w$, $f_j$, $f_r$), ensuring accurate reconstruction. Meanwhile, bounded joint amplitudes at transitional and out-of-distribution frequencies ($f_1$, $f_2$, $f_3$, $f_4$) confirm robust extrapolation capabilities.}

\label{fig:PNN_result}
\end{figure*}

\section{Experimental evaluation}



We evaluate the SPRINT framework through extensive simulations in Isaac Gym and physical deployments on the Unitree G1 platform. Specifically, the evaluation is structurally designed to address the following three questions:

\begin{itemize}
\item How do the spectral priors outperform conventional motion generative models, and how does this advantage manifest as enhanced tracking precision, training efficiency, and biomimetic naturalness?
\item How do the integral components of the SPRINT framework (i.e., prior guidance, residual control, progressive curriculum learning, and the AAC architecture) synergize to guarantee high-speed locomotion?
\item Does the SPRINT policy demonstrate zero-shot sim-to-real transfer during sprints and gait transitions?
\end{itemize}

\subsection{Experimental Setup}

\textbf{\textit{1) Simulation Environment}}.
We conduct highly parallelized policy training within the NVIDIA Isaac Gym environment \cite{makoviychuk2021isaac}, operating at a physics simulation frequency of 200\,Hz and a control update rate of 50\,Hz. The policy network architecture comprises a Multilayer Perceptron (MLP) with 256 hidden units for state feature extraction, coupled with a 64-unit Long Short-Term Memory (LSTM) module to capture complex temporal dynamics. To stabilize early-stage learning, we employ a progressive curriculum that gradually expands the velocity command sampling space from static standing to sprints. Specifically, this curriculum scales the target commands up to maximum bounds of $v_{x}^{\text{cmd}} \in [0, 6]$\,m/s (corresponding to $f \in [0.6, 2.3]$\,Hz) and $\omega_{z}^{\text{cmd}} \in [-0.7, 0.7]$\,rad/s. Leveraging the inherent efficiency of Isaac Gym's tensorized physics engine, the complete policy converges in approximately 6.5 hours on a single NVIDIA RTX 4090 GPU.

\textbf{\textit{2) Evaluation Metrics}}.
To rigorously validate the SPRINT framework, we distinguish the performance of the spectral generator from that of the final control policy. Accordingly, the evaluation metrics span three core dimensions: spectral prior fidelity ($L_{\text{rec}}$ and $E_{\text{BA}}$), biomimetic naturalness (FID) to evaluate both the prior and the policy, and tracking precision ($E_{\text{qpos}}$ and $E_{\text{vel}}$) to quantify the policy.

\begin{itemize}
\item \textbf{Reconstruction Loss} ($L_{\text{rec}}$)\textbf{:} This metric calculates the Mean Squared Error (MSE) between the joint trajectories $q$ generated by the spectral priors and the curated reference library $q_j^{\text{lib}}$ across the $n_{\text{p}}=5$ predefined frequencies for the joint subset $\mathcal{J}$ with $n_{\text{j}}=10$:

\begin{equation}
L_{\text{rec}} = \frac{1}{n_{\text{j}} n_{\text{p}}} \sum_{j \in \mathcal{J}} \sum_{f \in \mathcal{F}_{\text{p}}}  \left( q_{j}( f) - q_{j}^{\text{lib}}(f) \right)^2,
\end{equation}

where $\mathcal{F}_{\text{p}} = \{0.68, 0.86, 1.25, 1.36, 1.58\}$\,Hz denotes the primary frequency set of the reference library.

\item \textbf{Boundary Amplitude Error} ($E_{\text{BA}}$)\textbf{:} This metric evaluates the spectral priors' extrapolation reliability at the $n_{\text{e}}=2$ extreme frequencies ($\mathcal{F}_{\text{e}} = \{0.6, 2.3\}$\,Hz) relative to the reference boundaries $f_{\text{b}}$:

\begin{equation}
E_{\text{BA}} = \frac{1}{n_{\text{j}} n_{\text{e}}} \sum_{j \in \mathcal{J}} \sum_{f \in \mathcal{F}_{\text{e}}} \left| \text{Amp}_j(f) - \text{Amp}_j(f_{\text{b}}) \right|.
\end{equation}


\item \textbf{Fréchet Inception Distance} (FID)\textbf{:} We apply this metric at two levels: prior-level naturalness (comparing generated prior distributions against the reference library) and policy-level naturalness (measuring the discrepancy between spectral prior trajectories and actual joint trajectories). The FID\cite{heusel2017gans} evaluates the distributional similarity of motion sequences:

\begin{equation}
\text{FID} = \|\mu - \mu_w\|^2 + \text{tr}\left(\Sigma + \Sigma_w - 2(\Sigma\Sigma_w)^{\frac{1}{2}}\right),
\end{equation}
where $(\mu, \Sigma)$ and $(\mu_w, \Sigma_w)$ denote the statistics of the reference and target distributions, respectively. Lower values indicate superior morphological similarity.

\item \textbf{Imitation Error} ($E_{\text{qpos}}$)\textbf{:}  This metric computes the Mean Absolute Error (MAE) between the actual joint positions $q_j$ and the spectral reference $q_j^{\text{ref}}$:

\begin{equation}
E_{\text{qpos}} = \frac{1}{n_{\text{j}} n_{\text{c}}} \sum_{j \in \mathcal{J}} \sum_{v_{x}^{\text{cmd}} \in \mathcal{V}_{\text{c}}}  \left| q_{j}(v_{x}^{\text{cmd}}) - q_{j}^{\text{ref}}(v_{x}^{\text{cmd}}) \right| ,
\end{equation}

where $\mathcal{V}_{\text{c}}$ comprises commanded velocity sampled from $0$ to $6$\,m/s at $0.1$\,m/s increments, and $n_{\text{c}}$ denotes the number of elements in  $\mathcal{V}_{\text{c}}$.

\item \textbf{Velocity Tracking Error} ($E_{\text{vel}}$)\textbf{:} This metric measures the absolute deviation between the commanded longitudinal velocity $v_{x}^{\text{cmd}}$ and the actual velocity $v$:
\begin{equation}
E_{\text{vel}} = \frac{1}{n_{\text{c}} } \sum_{v_{x}^{\text{cmd}} \in \mathcal{V}_{\text{c}}}  \left| v_{x}^{\text{cmd}} - v \right|.
\end{equation}
\end{itemize}

\subsection{Comparative Analysis}

\renewcommand{\arraystretch}{1.5}
\setlength{\tabcolsep}{6pt} 
\begin{table}[h]
\centering
\caption{Quantitative Comparison of Motion Generators}
\begin{tabular}{@{}lccc@{}}
\hline
\noalign{\vskip-0.6pt}
\hline
Motion Generator & $L_{\text{rec}}$ ↓ & FID ↓ & $E_{\text{BA}}$ ↓ \\
\hline
AI-CPG~\cite{li2024ai}          & 0.0015          & 0.0042          & 0.7452          \\
\textbf{Spectral Priors}    & \textbf{0.0006} & \textbf{0.0008} & \textbf{0.0447} \\
\hline
\noalign{\vskip-0.6pt}
\hline
\end{tabular}
\label{tab:motion_generator}
\end{table}

\textbf{\textit{1) Spectral Priors Advantage}}.
The frequency-adaptive spectral priors outperform the AI-CPG baseline across tracking accuracy, biomimetic naturalness, and extrapolation reliability. As detailed in TABLE~\ref{tab:motion_generator}, our method achieves reductions in reconstruction loss, the FID, and boundary amplitude error. The first row of \figref{fig:PNN_result} visually highlights the intrinsic morphological adaptability of these priors. This adaptability facilitates direct deployment across diverse humanoid scales (including 1.7\,m, 1.3\,m and 1.1\,m) without necessitating platform-specific retargeting. Furthermore, as demonstrated in the second row of \figref{fig:PNN_result}, the near-perfect alignment between the joint trajectories generated by spectral priors and the reference library indicates that the priors maintain kinematic precision across the predefined frequencies (e.g., $f_w$, $f_j$, $f_r$). Ultimately, the nearly identical amplitudes confirm robust extrapolation to out-of-distribution frequencies. 

\begin{table*}[t] 
\centering
\caption{Quantitative Comparison with Baseline Methods}
\label{tab:method_comparison}
\renewcommand{\arraystretch}{1.5}
\setlength{\tabcolsep}{6pt} 

\begin{tabular}{@{}lccccccc@{}}
\hline
\noalign{\vskip-0.6pt}
\hline
\multirow{2}{*}{Method} & \multirow{2}{*}{Max $v_{\text{avg}}$ (m/s) ↑} & \multicolumn{3}{c}{Convergence Time (h) ↓} & \multirow{2}{*}{FID ↓} & \multirow{2}{*}{$E_{\text{qpos}}$ ↓} & \multirow{2}{*}{$E_{\text{vel}}$ ↓} \\
\cline{3-5}
& & $v_{x}^{\text{cmd}} \in [0, 4]$\,m/s & $v_{x}^{\text{cmd}} \in [0, 5]$\,m/s & $v_{x}^{\text{cmd}} \in [0, 6]$\,m/s & & & \\
\hline
Humanoid-Gym~\cite{gu2024humanoid} & 3.9048 & 4.5 & -- & -- & 2.3196 & 0.5283 & \textbf{0.0385} \\
AMP~\cite{peng2021amp}             & 4.7512 & 34.5 & 37 & -- & 1.7330 & 0.4652 & 0.1061 \\
AI-CPG~\cite{li2024ai}             & 4.8208 & 9 & 14 & -- & 0.8831 & 0.2277 & 0.0802 \\
\textbf{SPRINT}                    & \textbf{5.9648} & \textbf{3} & \textbf{4.5} & \textbf{6.5} & \textbf{0.8035} & \textbf{0.1868} & 0.0657 \\
\hline
\noalign{\vskip-0.6pt}
\hline
\end{tabular}
\end{table*}

\begin{figure}[htbp] 
\centering 
\includegraphics[width=\columnwidth]{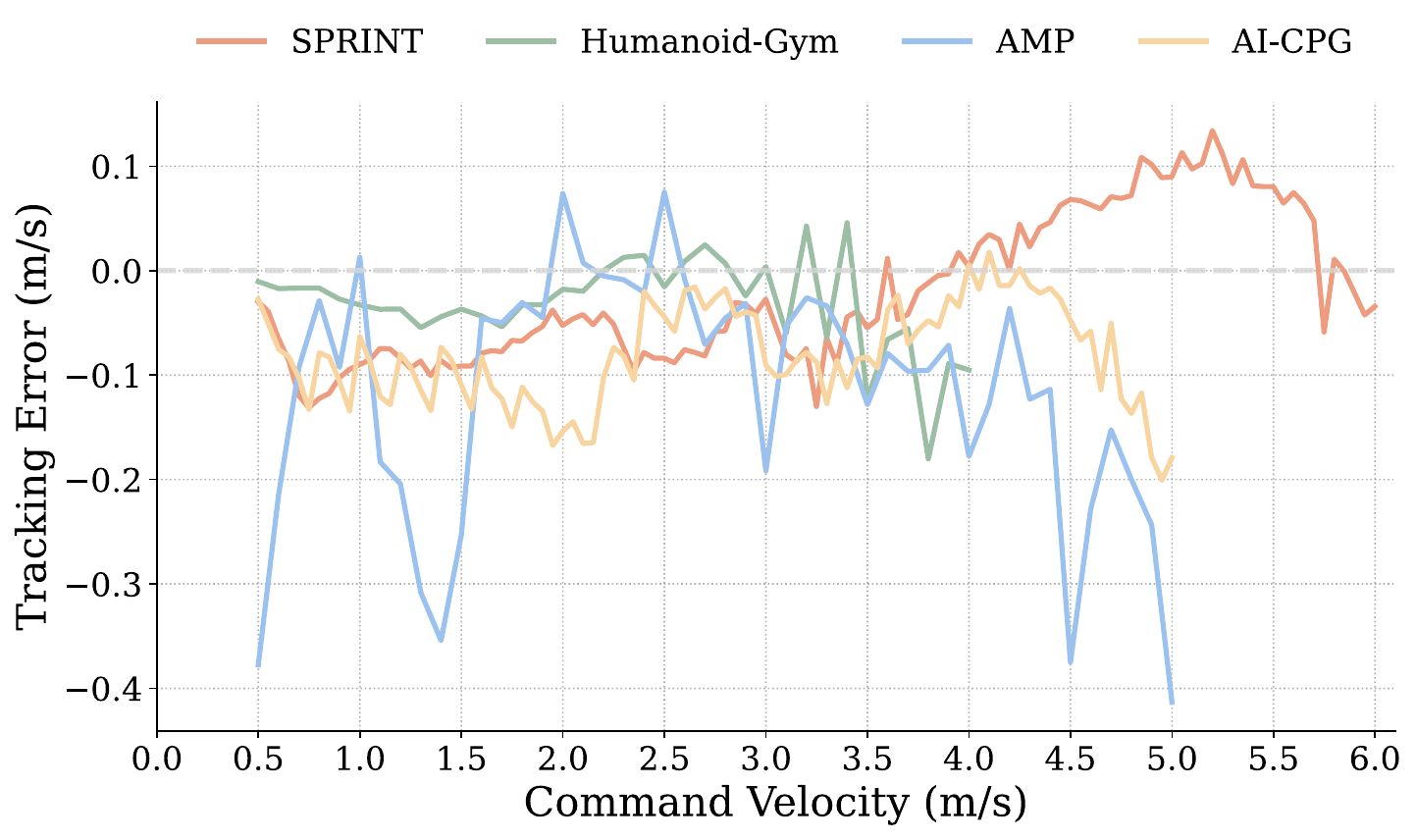}

\caption{Comparison of velocity tracking performance. While baseline frameworks exhibit low error at moderate speeds, they encounter destabilizing oscillations or experience failure during high-speed sprints. The SPRINT policy maintains bounded dynamic tracking precision across the entire continuous velocity spectrum, achieving stable locomotion at extreme velocities up to 6.0\,m/s.}


\vspace{-0.2cm} 
\label{fig:performance} 
\end{figure}

\textbf{\textit{2) SPRINT Policy Advantage}}.
SPRINT resolves the inherent trade-off between biomimetic naturalness and precise command execution during sprints. As evidenced by the quantitative metrics in TABLE~\ref{tab:method_comparison} and the tracking performance in \figref{fig:performance}, SPRINT achieves the highest average velocity while maintaining the lowest FID, imitation error and velocity tracking error. In contrast, RL frameworks like Humanoid-Gym~\cite{gu2024humanoid} impose rigid velocity constraints and yield unnatural movements, whereas AMP~\cite{peng2021amp} and AI-CPG~\cite{li2024ai} encounter motion instability during high-speed sprints and require prolonged optimization cycles. By constraining the RL exploration space using frequency-adaptive spectral priors and residual actions, SPRINT mitigates this instability. Furthermore, this restricted exploration accelerates learning, achieving stable locomotion at 4\,m/s in 3 hours and full convergence for a 6\,m/s sprint in approximately 6.5 hours. Ultimately, this rapid convergence and robust stability demonstrate the practical viability of SPRINT policy for seamless gait transitions and high-speed sprints.

\begin{figure*}[htbp]
\centering

\includegraphics[width=\textwidth]{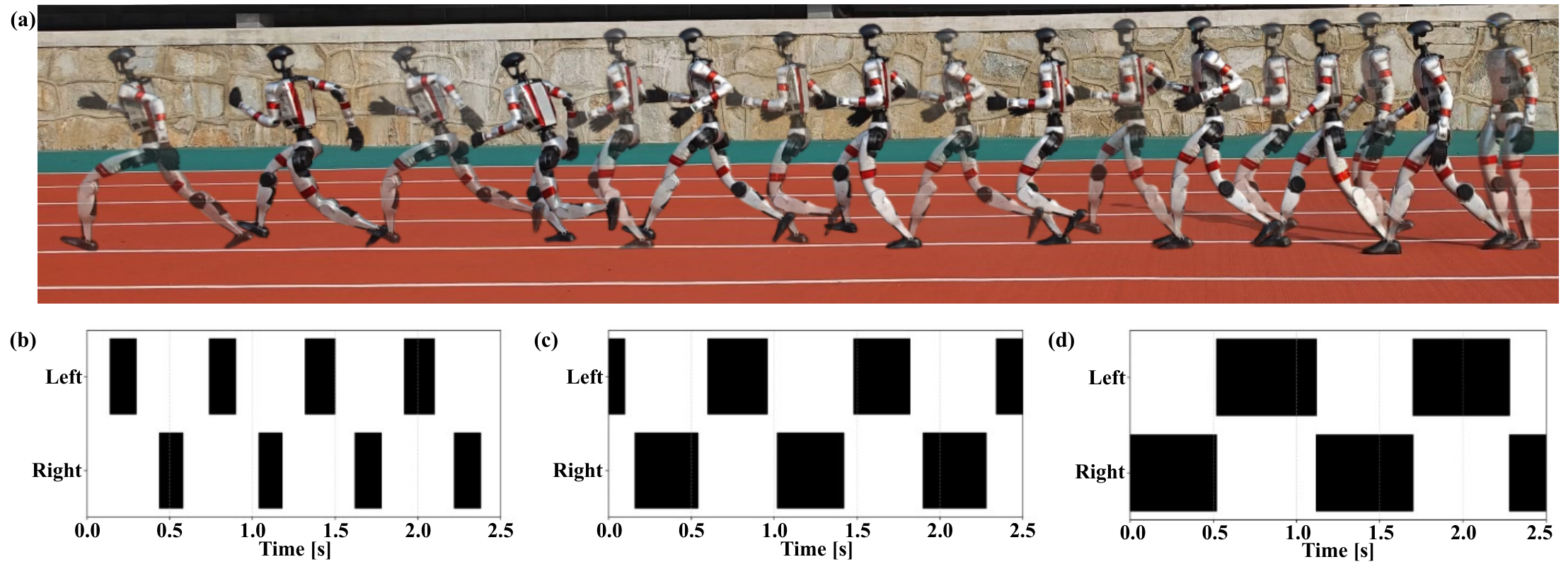}

\caption{Hardware validation of seamless gait transitions. \textbf{First row}: Time-lapse sequence capturing continuous acceleration on the Unitree G1 platform. \textbf{Second row}: Foot contact states confirming distinct locomotion patterns: (b) running and (c) jogging with clear flight phases, contrasted against (d) walking with a double-support phase.}

\label{fig:transition}
\end{figure*}

\subsection{Ablation Studies}
\renewcommand{\arraystretch}{1.5} 
\begin{table}[htbp]
\centering
\caption{Ablation Study Results}
\setlength{\tabcolsep}{1.5pt} 
\begin{tabular*}{\columnwidth}{@{\extracolsep{\fill}} @{}lcccc@{}}
\hline
\noalign{\vskip-0.6pt}
\hline
\text{Method} & Max $v_{\text{avg}}$ (m/s) ↑ & \text{FID} ↓ & $E_{\text{qpos}}$ ↓ & $E_{\text{vel}}$ ↓\\
\hline
SPRINT w/o $ r_{\text{prior}}$ & \textbf{6.0596} & 12.2305 & 1.0178 & 0.0932 \\
SPRINT w/o Curriculum        & 4.4810 & 1.3285  & 0.2336 & 0.1789 \\
SPRINT w/o Residual          & 5.9631 & 2.3285  & 0.4150 & 0.1179 \\
SPRINT w/o AAC                 & 5.4624 & 0.8777  & 0.1880 & 0.0755 \\
\textbf{SPRINT}              & 5.9648 & \textbf{0.8035} & \textbf{0.1868} & \textbf{0.0657} \\
\hline
\noalign{\vskip-0.6pt}
\hline
\vspace{-0.9cm} 
\end{tabular*}
\label{tab:ablation_study}
\end{table}

We conduct ablation experiments to rigorously quantify the individual contributions of SPRINT's core modules. By systematically ablating specific components, we evaluate their distinct impact on the overall dynamic performance, as summarized in TABLE~\ref{tab:ablation_study}.

\textbf{\textit{1) Effect without $r_{\text{prior}}$}}.
Ablating the spectral prior guidance reward degrades the overall motion quality, evidenced by an increased Imitation Error and FID, despite the robot achieving a peak velocity of 6.06\,m/s. Without the guidance of spectral priors, the agent optimizes purely for forward velocity, adopting unnatural movements to maximize speed. This confirms that frequency-adaptive spectral priors are crucial for maintaining biomimetic naturalness during sprints.

\textbf{\textit{2) Effect without Curriculum}}.
Removing the progressive curriculum destabilizes training and restricts the maximum achievable speed, which drops to 4.48\,m/s alongside a sharp rise in velocity tracking error. Directly exposing the agent to high-speed sprints from the onset creates an extremely challenging optimization landscape. Lacking the guidance of simpler tasks, the policy becomes trapped in suboptimal local minima, demonstrating that incremental task scaling is essential for acquiring stable sprinting behaviors.

\textbf{\textit{3) Effect without Residual}}.
Replacing residual control with absolute joint position targets reduces motion naturalness and tracking accuracy, reflected in elevated FID and tracking errors. This degradation occurs because predicting absolute positions directly expands the policy's exploration space. Conversely, SPRINT's residual architecture leverages spectral priors to strictly constrain exploration space, facilitating the stabilization required for athletic sprints.

\textbf{\textit{4) Effect without AAC}}.
Ablating the AAC architecture prevents the framework from achieving peak sprinting performance, evidenced by a drop in speed to 5.46\,m/s and reduced tracking precision. The absence of privileged environmental information impairs the critic network's value estimation, compromising its ability to effectively guide the actor network. This highlights the necessity of privileged state feedback for mastering high-speed locomotion skills.

\subsection{Hardware Experiment}
\begin{table}[htbp]
\centering
\caption{Sim-to-Real Transfer Performance on Unitree G1}
\label{tab:sim2real_transpose}
\renewcommand{\arraystretch}{1.15} 
\setlength{\tabcolsep}{5pt} 
\begin{tabular}{ccccccc} 
\toprule 
\multirow{2}{*}[-0.8ex]{\begin{tabular}[c]{@{}c@{}}$v_{x}^{\text{cmd}}$\\(m/s)\end{tabular}} & \multicolumn{2}{c}{FID} & \multicolumn{2}{c}{$E_{\text{qpos}}$} & \multicolumn{2}{c}{$v_{\text{avg}}$ (m/s)} \\
\cmidrule(lr){2-3} \cmidrule(lr){4-5} \cmidrule(lr){6-7} 
& Sim & Real & Sim & Real & Sim & Real\\
\midrule 
1 & 0.9931 & 1.0379 & 0.1912 & 0.1976 & 0.9090 & 0.9199 \\
2 & 0.7622 & 0.7867 & 0.1748 & 0.1852 & 1.9465 & 1.9031 \\
3 & 0.4991 & 0.4764 & 0.1564 & 0.1536 & 2.9717 & 2.9340 \\
4 & 0.6934 & 0.7342 & 0.1829 & 0.1885 & 4.0029 & 4.0606 \\
5 & 0.8920 & 0.8547 & 0.2154 & 0.2191 & 5.0887 & 5.0323 \\
6 & 1.1344 & 1.0970 & 0.2485 & 0.2510 & 5.9648 & 5.9143 \\
\midrule
\multicolumn{1}{l}{Mean Error} & \multicolumn{2}{c}{\textbf{0.0346}} & \multicolumn{2}{c}{\textbf{0.0052}} & \multicolumn{2}{c}{\textbf{0.0428}} \\

\bottomrule 
\end{tabular}
\end{table}

\textbf{\textit{1) Zero-Shot Sim-to-Real Transfer}}.
In TABLE~\ref{tab:sim2real_transpose}, real-world deployment results closely mirror the simulated metrics across speed, imitation accuracy, and biomimetic naturalness. This highly consistent performance across the continuous velocity spectrum is directly attributed to the synergy between the frequency-adaptive spectral priors and the hierarchical residual control architecture. By compensating for environmental disturbances, the SPRINT policy also guarantees the robust physical execution of sprints, as visually validated in \figref{fig:motivation}.

\textbf{\textit{2) Continuous Gait Transitions}}.
The robot utilizes a unified SPRINT policy to track continuous velocity commands and execute seamless gait transitions. As shown in the first row of \figref{fig:transition}, SPRINT policy enables smooth acceleration from static standing to a 4\,m/s running gait through intermediate walking and jogging states. Furthermore, foot contact states (second row of \figref{fig:transition}) reveal a transition from a double-support phase during walking to an aerial phase during jogging and running. Ultimately, these results validate SPRINT as an effective framework for enabling seamless gait transitions on physical humanoid robots.

\section{Conclusion}

The SPRINT framework employs frequency-adaptive spectral priors to achieve humanoid athletic sprints using a curated reference library of only five motion sequences. The spectral priors leverage the inherent periodicity of human locomotion to map discrete reference data onto a continuous kinematic manifold. This representation enables the generation of joint trajectories across a broad velocity spectrum, extrapolating to speeds that significantly exceed the reference distribution. Extensive field experiments on the Unitree G1 platform demonstrate that the SPRINT policy achieves a peak sprinting velocity of 6\,m/s while maintaining seamless gait transitions and high biomimetic naturalness. Ultimately, these results establish frequency-adaptive spectral priors as a highly efficient foundation for humanoid athletic sprints.

\bibliographystyle{ieeetr}
\bibliography{glorified,new_new}
\end{document}